\documentclass[final]{cvpr}

\usepackage{times}
\usepackage{epsfig}
\usepackage{multirow}
\usepackage{graphicx}
\usepackage{amsmath}
\usepackage{amssymb}
\usepackage{url}   
\usepackage{makecell}%
\usepackage{booktabs}       %
\usepackage{amsfonts}       %
\usepackage{nicefrac}       %
\usepackage{microtype}      %
\usepackage[dvispsnames]{xcolor}
\usepackage{graphicx}
\usepackage{comment}
\usepackage{amsmath,amssymb} %
\usepackage{pifont}
\usepackage{algorithmic}
\usepackage[ruled,vlined]{algorithm2e}
\usepackage{etoolbox}
\usepackage{subfigure}
\usepackage{floatrow}
\usepackage{wrapfig}
\usepackage{caption}
\newcommand{\atmostwidth}[1]{
\ifdim\width>#1
#1
\else
\width
\fi
}

\usepackage[pagebackref=true,breaklinks=true,colorlinks,bookmarks=false]{hyperref}
\usepackage{cleveref}

\definecolor{darkgreen}{RGB}{0, 100, 0}
\definecolor{purple}{RGB}{162, 50, 252}

\newtoggle{comments}
\togglefalse{comments}
\iftoggle{comments}{%
    \newcommand{\deva}[1]{{\leavevmode\color{blue}[Deva: #1]}}
    \newcommand{\achal}[1]{{\leavevmode\color{magenta}[Achal: #1]}}
    \newcommand{\tarasha}[1]{{\leavevmode\color{darkgreen}[Tarasha: #1]}}
    \newcommand{\yiting}[1]{{\leavevmode\color{magenta}[Yi-Ting: #1]}}
    \newcommand{\kangle}[1]{{\leavevmode\color{darkgreen}[Kangle: #1]}}
  \newcommand{\devapop}[1]{\todo[color=blue!40]{Deva: #1}}
  \newcommand{\achalpop}[1]{\todo[color=magenta!40]{Achal: #1}}
    \newcommand{\change}[1]{{\leavevmode\color{red}#1}}
    \newcommand{\newstuff}[1]{{\leavevmode\color{purple}#1}}
}{%
  \newcommand{\deva}[1]{}
  \newcommand{\achal}[1]{}
  \newcommand{\tarasha}[1]{}
  \newcommand{\yiting}[1]{}
  \newcommand{\kangle}[1]{}
  \newcommand{\devapop}[1]{}
  \newcommand{\achalpop}[1]{}
  \newcommand{\change}[1]{#1}
  \newcommand{\newstuff}[1]{#1}
}

\newcommand{\cmark}{\ding{51}}%
\newcommand{\xmark}{\ding{55}}%

\renewcommand{\t}{\mathcal{T}}
\renewcommand{\d}{\mathcal{D}}

\SetKw{KwIn}{in}
\SetKw{KwData}{Data:}

\def\cvprPaperID{4348} %
\def\confYear{CVPR 2021}

\begin{document}

\title{Detecting Invisible People}

\author{
Tarasha Khurana$^1$ \hspace{5mm} Achal Dave$^1$ \hspace{5mm} Deva Ramanan$^{1, 2}$\vspace{2mm}\\
$^1$Carnegie Mellon University \hspace{5mm} $^2$Argo AI\\
{\tt\small \{tkhurana, achald, deva\}@cs.cmu.edu}
\vspace{3mm}
}

\twocolumn[{%
\vspace{-1em}
\maketitle
\vspace{-1em}
\centering
\includegraphics[width=\linewidth]{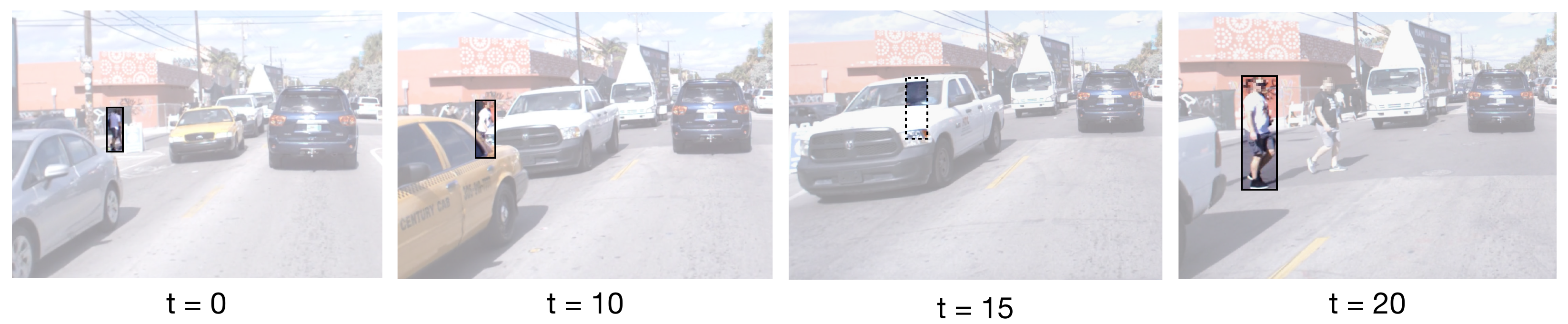}\\
\captionof{figure}{
  We visualize an online tracking scenario from Argoverse~\cite{chang2019argoverse} that requires tracking a pedestrian through a complete occlusion. Such applications cannot wait for objects to re-appear (\eg, as re-identification approaches do): autonomous agents must properly react \textit{during} the occlusion. We treat online detection of occluded people as a {\em short-term forecasting} challenge.\vspace{1em}}
\label{fig:splash}
\vspace{6mm}
}]

\begin{abstract}
  \vspace{-3mm}
   Monocular object detection and tracking have improved drastically in recent years, but rely on a key assumption: that objects are visible to the camera. Many offline tracking approaches reason about occluded objects {\em post-hoc}, by linking together tracklets after the object re-appears, making use of reidentification (ReID). However, online tracking in embodied robotic agents (such as a self-driving vehicle) fundamentally requires object permanence, which is the ability to reason about occluded objects {\em before} they re-appear. In this work, we re-purpose tracking benchmarks and propose new metrics for the task of detecting invisible objects, focusing on the illustrative case of people. We demonstrate that current detection and tracking systems perform dramatically worse on this task. We introduce two key innovations to recover much of this performance drop. We treat occluded object detection in temporal sequences as a short-term forecasting challenge, bringing to bear tools from dynamic sequence prediction.
  Second, we build dynamic models that explicitly reason in 3D, making use of observations produced by state-of-the-art monocular depth estimation networks.
  To our knowledge, ours is the first work to demonstrate the effectiveness of monocular depth estimation for the task of tracking and detecting occluded objects. Our approach strongly improves by 11.4\% over the baseline in ablations and by 5.0\% over the state-of-the-art in F1 score. 
\end{abstract}

\section{Introduction}

Object detection has seen immense progress, albeit under a seemingly harmless assumption: that objects are \textit{visible to the camera} in the image.
However, objects that are fully occluded (and thus, invisible) continue to exist and operate in the world. Indeed, object permanence is a fundamental visual cue exhibited by infants in as early as 3 months~\cite{baillargeon1991object,huang2005tracking}. Practical autonomous vision systems must similarly reason about objects under such occlusions in order to ensure safe operation (Figure~\ref{fig:splash}). Interestingly, existing work on object detection and tracking tends to de-emphasize this capability, either choosing to completely ignore highly-occluded instances for evaluation \cite{everingham2010pascal, lin2014microsoft, russakovsky2015imagenet, xu2018youtube}, or simply downweighting them because they occur so rarely that they fail to materially affect overall performance \cite{milan2016mot16}. One reason that invisible-object detection may have been under-emphasized in the tracking community is that for {\em offline} analysis, one can post-hoc reason about the presence of an occluded object by relinking detections {\em after} it reappears. This approach has spawned the large subfield of reidentification (ReID). However, in an {\em online} setting (such as an autonomous vehicle that must make decisions given the available sensor information), intelligent agents must be able to instantaneously reason about occluded objects {\em before} they re-appear.

{\bf Problem formulation:} We begin by introducing \newstuff{benchmarks and metrics} for evaluating the task of detecting and tracking invisible people.
To do so, we repurpose annotations from existing tracking \newstuff{benchmarks} and introduce \newstuff{metrics} for evaluating this task that appropriately rewards detection of occluded people. To ensure \newstuff{benchmarks are} online, we forbid algorithms from accessing future frames when reporting object states for the current frame. Although this task requires reasoning about object trajectories, it can be evaluated as both a \textit{detection} and a \textit{tracking} problem. For the latter, we introduce extensions to tracking metrics. When analyzing our metrics, it becomes readily apparent that human annotation of occluded objects is challenging. We provide pilot human vision experiments in Section \ref{sec:experiments} that show annotators are still consistent, but exhibit larger variation in labeling the pixel position of occluded instances. This suggests that algorithms for occluded object detection should report {\em distributions} over object locations rather than precise discrete (bounding box) locations. 
Inspired by metrics for evaluating multimodal distributions in the forecasting literature~\cite{chang2019argoverse}, we explore probabilistic algorithms that make $k$ predictions which are evaluated by Top-$k$ accuracy.

{\bf Analysis:} Perhaps not surprisingly, our first observation is that performance of state-of-the-art detectors and trackers plummets on occluded people,
from 68.5\% to 28.4\%; it is far easier to detect visible objects than invisible ones!
This underscores the need for the community to focus on this underexplored problem.
We introduce two simple but key innovations for addressing this task, which collectively improve performance from 28.4\% to 39.8\%. (a) We recast the problem of online tracking of occluded objects as a {\em short-term forecasting} challenge. We explore state-of-the-art (SOTA) deep forecasting networks, but find that classic linear dynamic models (Kalman filters) perform quite well. (b) Because modeling occlusions is of central importance, we cast the problem as one of 3D tracking given 2D image measurements. While there exists considerable classic work in this direction \cite{roach1980determining, broida1990recursive, spinello2010nonlinear, civera2008inverse}, we make use of SOTA {\em monocular depth estimation} networks that infer depth from 2D images. While these do not provide metric-accurate depth, we find that they produce ``good enough'' estimates of relative depth, allowing our dynamic models to reason about occlusions arising from relative depth orderings and freespace constraints. To our knowledge, ours is the first work to demonstrate the effectiveness of monocular depth estimation for tracking and detecting occluded objects.

{\bf Overview:} After reviewing related work, we present our core algorithmic contributions, which include straightforward but crucial extentions to classic (Kalman) linear dynamic models that allow them to (a) take advantage of putative depth observations from a monocular network and (b) forecast object state even during occlusions. Finally, we conclude with an extensive evaluation on \newstuff{three datasets \cite{milan2016mot16, wang2020panda, dendorfer2020mot20} repurposed for detecting occluded objects.}

\section{Related Work}

\textbf{Amodal object detection} aims to segment the full extent of objects that may be \textit{partially} (but not fully) occluded.
\cite{zhu2015semantic} introduces the task of amodal semantic segmentation with a dataset labeled by multiple annotators, which is later expanded by~\cite{zhu2017semantic}. 
More recently, \cite{qi2019amodal} introduces a larger dataset of amodal annotations on the KITTI~\cite{geiger2013vision} dataset.
Approaches that tackle these tasks largely rely on training variants of standard detection models (e.g., \cite{he2017mask}) on amodal annotations that are synthetically generated from modal datasets~\cite{li2016amodal,ehsani2018segan,zhang2019learning,yan2019visualizing}.
As this line of work addresses object detection from a single image, it focuses only on objects that are at least \textit{partially visible}.
By contrast, we target fully occluded people, which cannot be recovered from a single frame.

\textbf{Multi-object tracking} requires tracking across partial and full occlusions.
Approaches for this task address occlusions post-hoc in an \textit{offline} manner, using appearance-based re-identification models to identify occluded objects after they become visible.
These appearance-based models can be incorporated into tracking approaches, as part of a graph optimization problem \cite{berclaz2011multiple,pirsiavash2011globally,yu2007multiple} or online linking \cite{Wojke2018deep,bergmann2019tracking}.
In this work, we point out that some approaches \textit{internally} maintain online estimates of the position of occluded people~\cite{bergmann2019tracking,Bewley2016_sort,Wojke2018deep}, but explicitly choose not to report these internal predictions, as they tend to be noisy and, thus, are penalized heavily by current benchmarks.
We provide two simple extensions to these internal predictions that significantly improve detection of occluded people while preserving accuracy on visible people. 
\newstuff{\cite{grabner2010tracking} tracks occluded objects using contextual `supporters', but requires a user to initialize a single object to track in uncluttered scenes; by contrast, we simultaneously detect and track people in large crowds.}
Finally, many surveillance-based tracking systems explicity reason about object occupancy with respect to ground plane coordinates (computed through a homography~\cite{fleuret2007multicamera}), often using multiple cameras to track through occlusions~\cite{khan2006multiview,kim2006multi}. We focus on the monocular case where the camera may move.

\textbf{Forecasting} approaches predict pedestrian trajectories in future, unobserved frames.
These approaches leverage social cues from nearby pedestrians or semantic scene information to better model person trajectories~\cite{scovanner2009learning,leal2011everybody,yamaguchi2011you,pellegrini2009you,ma2017forecasting,kitani2012activity}.
Recently, data-driven approaches have also been proposed for learning social cues~\cite{alahi2016social,robicquet2016learning}.
We note that detection of fully occluded people can be formulated as forecasting the trajectory of a visible person in future frames, where the positions of the occluded person are unobserved, but the rest of the frame \textit{can} be observed.
Our approach uses a constant-velocity model to forecast trajectories, equipped with depth cues from the observed frames, to improve detection of occluded people.
In Section \ref{sec:ablation}, we show that while this approach can use a more powerful forecasting model, the constant-velocity approximation is sufficient in our setting.

\section{Method}

We build an online approach for detecting invisible people starting with a simple tracker, using estimated trajectories of visible people to forecast their location during occlusions. We describe our tracking mechanism, building upon~\cite{wojke2017simple}.
While such trackers \textit{internally} forecast the location of occluded people for improved tracking, these forecasts tend to be noisy and cannot directly localize occluded people.
To address this, we incorporate depth cues from a monocular depth estimator to reason about occlusions in 3D.

\subsection{Background}
\label{sec:background}
To detect people during occlusions, we build on a simple online tracker~\cite{wojke2017simple} that estimates the trajectories of visible people. We briefly describe aspects relevant to our approach, but refer the reader to~\cite{wojke2017simple} for a more detailed explanation.
In the first frame, this tracker instantiates a track for each detected person.
The tracker adds each track to its ``active'' set, representing people that have been seen so far.
Each track maintains a Kalman Filter whose state space encodes the position ($x$, $y$), aspect ratio ($a$), height ($h$), and corresponding velocities ($\dot{x}, \dot{y}, \dot{a}, \dot{h}$) of the person.%
The filter's process model assumes a constant velocity model with gaussian noise (i.e., $x_t = x_{t-1} + \Dot{x_{t-1}} + \epsilon_x$).
At each successive frame, the tracker first runs the \textit{predict} step of the filter, using the process model to forecast the location of the track in the new frame.
Next, each detection in the current frame is matched to this set of active tracks based on appearance features, and distance to the tracks' forecasted location (as estimated by the filter). 
A new track is created for all detections that are unmatched.
If a track is matched to a detection, the detection is used as a new observation to update the track's filter, and the detection is reported as part of the track.
Importantly, if a track does not match to any detection, its forecasted box is \textit{not} reported. When a track is not matched to a detection for more than $N_{\textrm{age}}$ frames, it is deleted.

\subsection{Short-term forecasting across occlusions} 
Although this tracker \textit{internally} forecasts the positions of all tracks at each step, its estimates are used only to improve the association of tracks to detections, and are not reported externally.
However, these internally forecasted track locations are crucial as they may correspond to an occluded person.
We show that naively reporting these track locations leads to significant \textit{recall} of occluded people, but the noise in these estimates results in poor precision.
Further, these noisy estimates lead to a small decrease in \textit{overall} accuracy, as standard benchmarks largely focus on visible people.
We improve these estimates by augmenting them with  3D information.
Specifically, we use a monocular depth estimator \cite{li2018megadepth} to get per pixel depth estimates of the scene.
We then augment our Kalman Filter state space with the \textit{inverse} depth.
Inverse depth is a commonly used representation predicted by depth estimators \cite{li2018megadepth, lasinger2019towards} due to important benefits, including the ability to represent points as infinity and ability to model uncertainty in pixel disparity space (commonly used for stereo-based depth estimation~\cite{okutomi1991multiple}).
Our state space thus additionally includes $1/z$ variable. 

\begin{figure*}[t]
  \centering
  \includegraphics[width=0.95\linewidth]{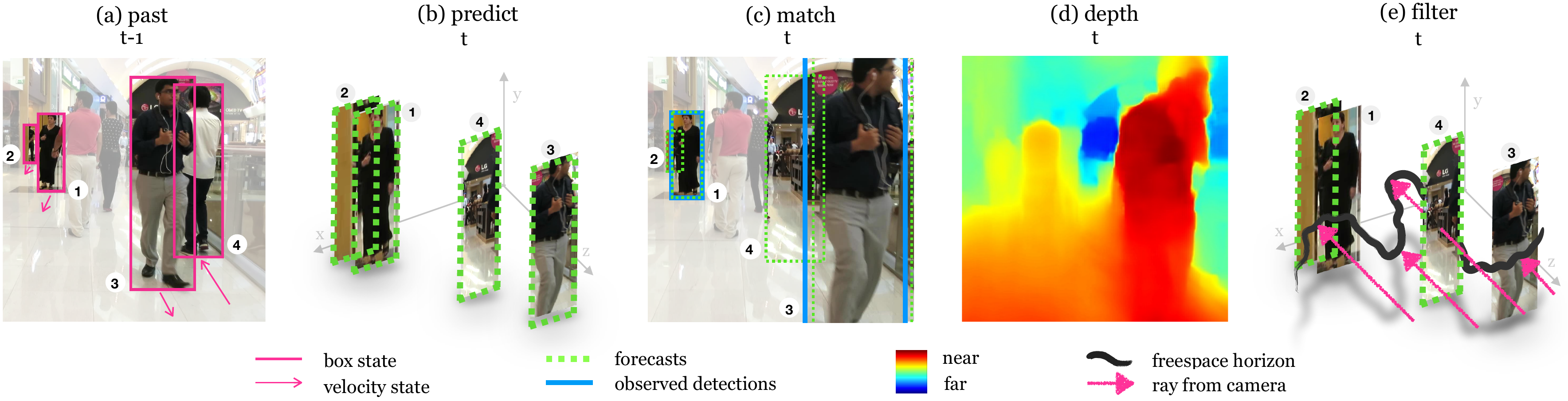}
  \caption{
 (a) Frame $t-1$ has active tracks $\{1, 2, 3, 4\}$, each with an internal state of its 2D position, size, velocity, and \textit{depth} (see text). (b) We forecast tracks in 3D for frame $t$. %
 (c) Tracks are matched to observed detections at $t$ using spatial and appearance cues.
 Matched tracks are considered visible (\eg 1, 3).
 Tracks which don't match to a visible detection (\eg 2, 4) may be occluded, or simply incorrectly forecasted.
 (d) To resolve this ambiguity, we leverage depth cues from a monocular depth estimator, 
 to compute (e) the \textit{freespace horizon}. The region between the camera and the horizon must be freespace, while the area beyond it is unobserved, and so may contain \textit{occluded} objects.
 Tracks lying beyond the freespace horizon  are reported as occluded (\eg 2).
 Tracks \textit{within} freespace (\eg 4) should have been visible, but did not match to any visible detections. Hence, we assume these tracks are incorrectly forecasted, and we delete them. 
  \vspace{-3mm}
  }
  \label{fig:methodtk}
\end{figure*}

\subsection{Tracking in 3D camera coordinates using 2D image coordinates}
\label{sec:temporalnoise}

Equipped with depth estimates, we formulate tracking with a constant velocity assumption in 3D using 2D measurements.
\change{We make simplifying assumptions here for exposition, but show that our method works even when these are relaxed.}
Concretely, let us model objects as cylinders with centroids $(X_t,Y_t,Z_t)$, height $H$ and aspect ratio $A_t$. We model object height as constant, but allow for varying aspect ratios because people %
are non-rigid.
In order to simplify notation, assume pinhole optics with a known focal length $f$. %
We can then compute image-measured bounding boxes with centroid $(x_t,y_t)$ and dimensions $(h_t,a_t)$ as follows:
\begin{align}
    x_t = f\frac{X_t}{Z_t}, \quad y_t = f\frac{Y_t}{Z_t}, \quad h_t = f\frac{H}{Z_t}, \quad a_t = {A_t}
\end{align}

Let us assume a constant velocity motion model in 3D with Gaussian noise:
\begin{align}
        X_t &= X_{t-1} + \dot{X}_{t-1} + \epsilon_X, \quad \epsilon_X \sim N(0,\sigma_X), \label{eq:motion}
\end{align}
where similar equations hold for $Y_t$, $Z_t$ and $A_t$. Assume image measurements are given by perspective projection followed by Gaussian image noise and the observed (inverse) depth from a depth estimator associated with an object is $1/z_t$. This results in the following projection equations:
\begin{align}
        x_t = f\frac{X_t}{Z_t} + \epsilon_x, \quad \epsilon_x \sim N(0,\sigma_x) \\
        \frac{1}{z_t} = \frac{1}{Z_t} + \epsilon_z, \quad \epsilon_z \sim N(0,\sigma_z) \label{eq:project}
\end{align}
\noindent with similar equations for $y_t$, $h_t$, and  $a_t$. 
Note that inverse depth naturally assumes a large uncertainty in far away regions, and a small uncertainty in nearby regions.
Defining a 3D state space leads us to a modified formulation, written as 
\begin{equation}
    \left(f\frac{X_t}{Z_t},f\frac{Y_t}{Z_t},\frac{1}{Z_t},A_t,f\frac{H}{Z_t},f\frac{\dot{X}_t}{Z_t},f\frac{\dot{Y}_t}{Z_t}, \dot{A_t}\right)
\end{equation}
We can therefore rewrite \Cref{eq:motion} as:
\begin{align}
        f\frac{X_t}{Z_{t}} &\approx f\frac{X_t}{Z_{t-1}} = f\frac{X_{t-1}}{Z_{t-1}} + f\frac{\dot{X}_{t-1}}{Z_{t-1}} + f\frac{\epsilon_X}{Z_{t-1}}\\
        x_t &\approx x_{t-1} + \dot{x}_{t-1} + f\frac{\epsilon_X}{Z_{t-1}}
\end{align}
\noindent where the approximation holds if depths are smooth over time $(Z_t \approx Z_{t-1})$. Technically, the above is no longer a linear dynamics model since the noise depends on the state. But the equation suggests that {\em one can approximately apply a Kalman filter on 2D image measurements augmented with a temporal noise model that is scaled by the estimated inverse-depth of the object}.
Intuitively, this suggests that one should enforce smoother tracks for objects far away. Our approach thus scales the process noise ($\epsilon_X$) for far away objects, leading to more accurate predictions. Algorithmically, \cite{wojke2017simple} by default scales process and observation noise covariances according to the person’s height; our approach instead multiplies the process covariance by the person's estimated depth, computed by aggregating past monocular depth observations %
and state estimates over time.

\change{\textbf{Relaxing assumptions.}
The derivation above relies on three simplifying assumptions.
First, we assumed the camera focal length $f$ was known. In many practical applications, it is possible to calibrate the camera so that this assumption is satisfied.
However, we evaluate on video sequences from datasets where no camera intrinsics are provided.
Instead of calculating $f$, we directly tune the $f$-scaled variances (e.g., $f\sigma_X$) on the train set.
We make two additional assumptions: that people move with constant velocity in 3D, and that depth estimates are smooth over time.
Although these assumptions do not always hold in real world scenarios, we empirically find that our method generalizes to diverse scenarios.
}

\textbf{Filtering estimates lying in freespace.}
Equipping our state space with depth information allows us to forecast 3D trajectories.
Meanwhile, applying a monocular depth estimator allows us to determine regions in 3D space that are occluded to the camera.
Specifically, if our approach forecasts a person at a point $P_f = (x_f, y_f, z_f)$, we can determine whether $P_f$ should be visible to the camera by estimating whether $P_f$ lies in the freespace between the camera and its nearest occluder.
\newstuff{In the filter stage} in Figure \ref{fig:methodtk}, we visualize one slice of the ``freespace horizon'': points beyond this horizon are occluded, while points between the camera and the horizon should be visible.

Concretely, let $z_o$ be the (observed) depth of the horizon at $(x_f, y_f)$.
If the forecasted depth ($z_f$) lies closer to the camera than the horizon depth ($z_o$), as with person ``4'' in Figure \ref{fig:methodtk} (e), then the person must be in the \textit{freespace} between the camera and its closest object, and therefore visible.
If we \textit{do not} detect this person, then we assume the forecast is an error, and either suppress the forecasted box for the current frame (in the case of small errors, when $z_f < \alpha_{\textrm{supp}} z_o$) or delete the track entirely (for large errors, when $z_f < \alpha_{\textrm{delete}} z_o$).
A key advantage of this approach is the ability to reason about occlusions arising not only from interactions between tracked people, but also from natural occluders such as trees or cars. 
Section~\ref{sec:ablation} shows that this modification is critical for improving the precision of our trajectory forecasts.

\textbf{Camera motion.}
Camera motion is challenging, as our approach assumes linear dynamics for trajectories.
To address this, we follow prior work (e.g., \cite{bergmann2019tracking}) in estimating a non-linear pixel warp $W$ between neighboring frames which maps pixel coordinates ($x_{t-1}, y_{t-1}$) in one frame to the next ($x_t, y_t$).
This warp is then used to align boxes forecasted using frames up to $t-1$ with frame $t$.
Note that this alignment assumes the motion of dynamic objects is small relative to the scene motion, allowing for the use of an image registration algorithm \cite{evangelidis2008parametric}.
Despite the simplicity of this modification, we show \newstuff{in the appendix} that it helps considerably for the moving camera sequences. 
We also detail our algorithm with pseudo-code in the appendix. We proceed to an empirical analysis of the task and prior methods, showing the benefits of each component of our proposed approach.

\section{Experimental Results}
\label{sec:experiments}
We first describe our proposed \newstuff{benchmarks}, including the datasets and our proposed \newstuff{metrics} for evaluating the task of detecting occluded people.
Next, we conduct an oracle study in Section~\ref{sec:oracle} to analyze how well existing approaches can detect occluded people. We then compare our proposed approach to these state-of-the-art approaches in multiple settings in Section~\ref{sec:sota}. Finally, we analyze each component of our approach with a detailed ablation study in Section~\ref{sec:ablation}.

\textbf{Dataset.} %
Evaluating our approach is challenging, as most datasets do not annotate occluded objects.
The MOT-17 \cite{milan2016mot16}\newstuff{, MOT-20 \cite{dendorfer2020mot20} and PANDA \cite{wang2020panda} datasets} are key exceptions which label both visible and occluded people, along with a \textit{visibility} field indicating what portion of the person is visible to the camera.
We find that a majority of the annotations in these datasets (over 85\% in each dataset) are people that are at least partially visible, leading standard evaluations on these datasets to underemphasize occluded people.
To address this, we separately evaluate accuracy on the subset of fully \textit{occluded} people (indicated by $<10\%$ visibility).
\newstuff{MOT-17 contains 7 sequences with publicly available groundtruth, and 7 test sequences with held-out groundtruth. 
We evaluate on these 14 sequences. MOT-20 contains 8 sequences, of which 4 have held-out groundtruth. PANDA officially releases a high-resolution 2FPS groundtruth for its 10 train and 5 test sequences. Because tracking and forecasting is challenging at such low frame rates, we reached out to the authors who provided a high-frame rate (30FPS), low-resolution groundtruth for 9 train videos. We report results on MOT-20 and PANDA train set without tuning our pipeline on any of the videos in these datasets. From visual inspection, we found that visibility labels in PANDA tend to be noisy (see the appendix), and so we define objects with up to 33\% visibility as occluded.
We carry out the analysis including oracle and ablation study on MOT-17 train and report the final results on MOT-17 test, MOT-20 and PANDA datasets. In all, these three datasets target a diverse set of application scenarios -- static surveillance cameras, car-mounted cameras, and hand-held cameras.} 

\begin{figure}[t]
\centering
\includegraphics[height=7em]{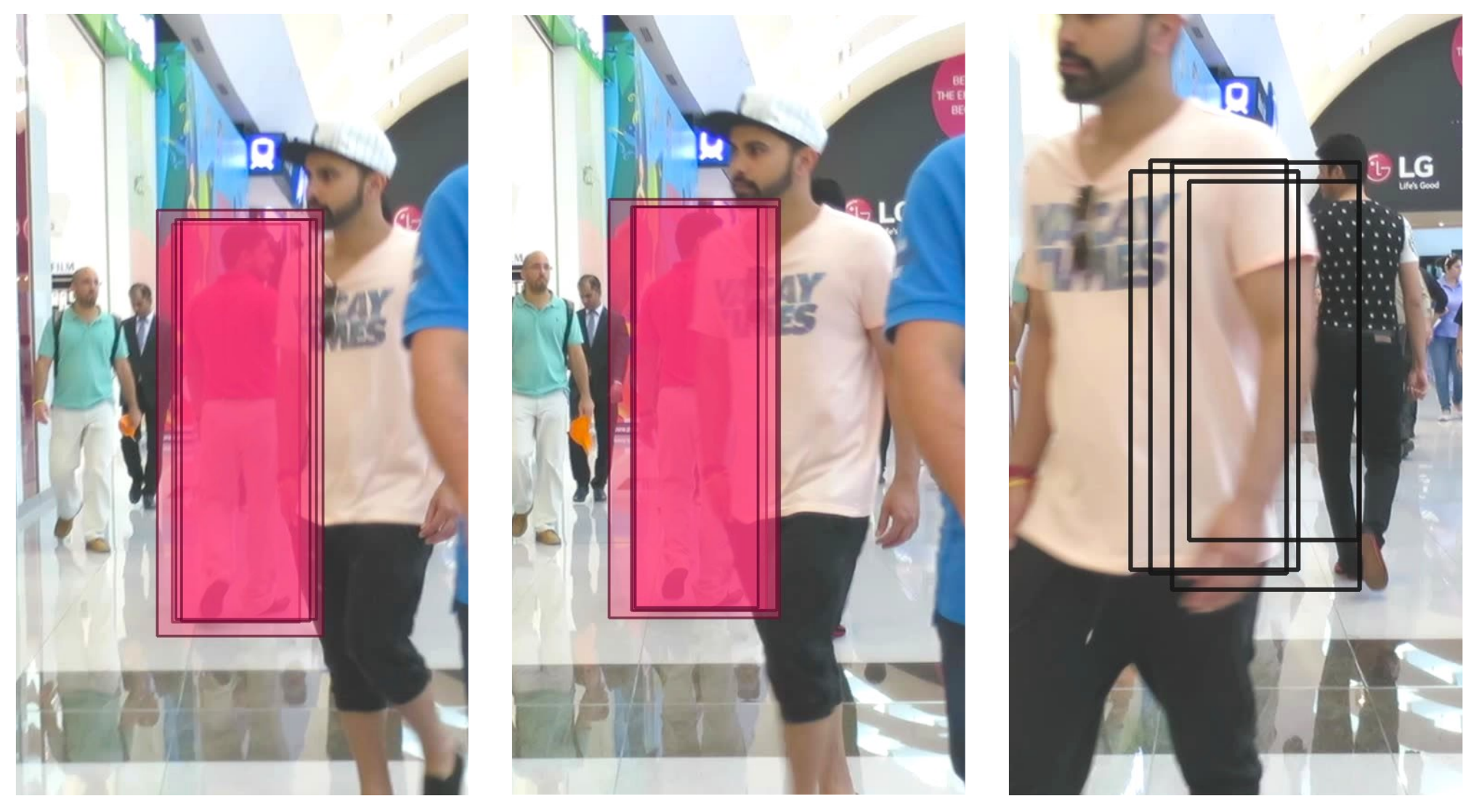}
\hspace{0pt}
\includegraphics[height=7em]
{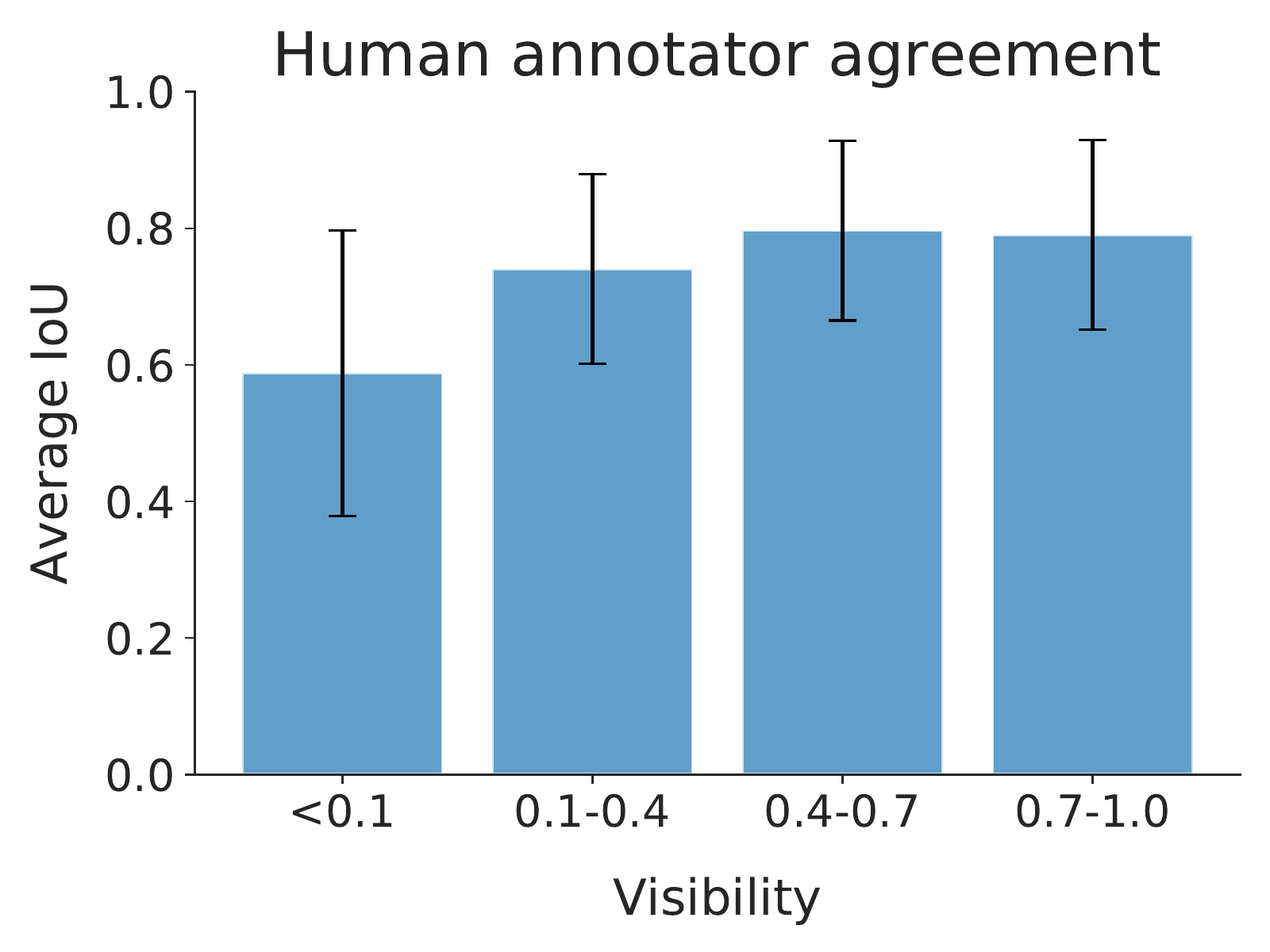}
\caption{We visualize bounding boxes labeled by multiple (4) in-house annotators ({\bf left}). During small occlusions, annotators strongly agree. During large occlusions (less than 10\% visible, last frame), annotators still agree to a fair extent (average IoU overlap of 60\%, {\bf right}), but require temporal video context. We use these to justify our Top-$k$ evaluation and motivate our probabilistic tracking approach.
\vspace{-5mm}} 
\label{fig:human_distribution}
\end{figure}

\renewcommand{\tabcolsep}{0.15cm}
\begin{table*}[t]
\caption{Oracle ablations on MOT-17 train reporting Top-5 F1, Top-1 F1 and IDF1 for occluded and all people, using Faster R-CNN detections. `Occl strat' stands for Occlusion Strategy. We report the Top-5 mean and standard deviation for 3 runs. 
\vspace{-2mm}
}
\centering
 \resizebox{0.95\linewidth}{!}{
\begin{tabular}{lccccccccccc}
\toprule[1.5pt]
 \multirow{2}{*}{Detections} & \multirow{2}{*}{Tracks} &
\multirow{2}{*}{Occl Strat} &
\multirow{2}{*}{Online?} & \multicolumn{4}{c}{Top-5} & \multicolumn{2}{c}{Top-1 F1} & \multicolumn{2}{c}{IDF1} \\
 \cmidrule(lr){5-8}
\cmidrule(lr){9-10}
\cmidrule(lr){11-12}
&&& & Occl F1 & Occl Prec & Occl Rec & All F1 & Occl & All & Occl & All\\

\midrule[.5pt]
Groundtruth (vis.) & Groundtruth & Interpolate & \xmark  & 87.3 {\scriptsize$\pm$0.1} & 83.8 {\scriptsize$\pm$0.2} & 91.1 {\scriptsize$\pm$0.1} & 98.0 {\scriptsize$\pm$0.0} & 79.8 & 96.8 & 77.8 & 96.7\\
Faster R-CNN & Groundtruth & Interpolate & \xmark  & 46.4 {\scriptsize$\pm$0.1}  & 65.5 {\scriptsize$\pm$0.1} &  35.9 {\scriptsize$\pm$0.1} & 70.5 {\scriptsize$\pm$0.0} & 34.4 & 68.1 & 20.5 & 67.4\\
\midrule[.5pt]
Groundtruth (vis.) & DeepSORT & Interpolate & \xmark & 53.3 {\scriptsize$\pm$0.2} & 86.7 {\scriptsize$\pm$0.1} & 38.5 {\scriptsize$\pm$0.2} & 92.3 {\scriptsize$\pm$0.0} & 44.4 & 92.0 & 21.3 & 81.0\\
Faster R-CNN & DeepSORT & Interpolate & \xmark & 32.2 {\scriptsize$\pm$0.0} & 60.8 {\scriptsize$\pm$0.2} & 21.9 {\scriptsize$\pm$0.0} & 69.9 {\scriptsize$\pm$0.0} & 23.2 & 68.4 & 6.4 & 53.3\\
\midrule[.5pt]
Faster R-CNN & DeepSORT & Forecast & \cmark & 29.8 {\scriptsize$\pm$0.2} & 29.5 {\scriptsize$\pm$0.4} & 30.2 {\scriptsize$\pm$0.1} & 69.4 {\scriptsize$\pm$0.0} & 20.9 & 66.5 & 7.6 & 53.3\\
\bottomrule[1.5pt]
\end{tabular}}
\label{tab:oracle_simple}
\end{table*}

\textbf{Metric.}
\newstuff{
As most benchmarks consist primarily of visible people, existing metrics which measure performance across all people underemphasize the accuracy of detecting occluded people.
We propose detection and tracking metrics which evaluate accuracy on occluded people, as indicated by visibility $<10\%$ and on all (visible and invisible) people.
Since localizing fully-occluded people involves higher positional uncertainty than visible people, we allow algorithms to predict $k$ potential locations for each person. 

\textbf{Top-$k$ F1:} We start by modifying the standard detection evaluation protocol \cite{everingham2010pascal, lin2014microsoft}. For every person, we allow methods to report $k$ predictions, $P=\{p_1, p_2, \dots , p_k\}$.
We match these predictions to all groundtruth boxes based on intersection-over-union (IoU).
We define the overlap between a groundtruth $g$ and $P$ as the maximum overlap with the predictions $p_i$ in $P$ --- \ie, $\text{IoU}(g, P) = \text{max}_i \text{IoU}(g, p_i)$.  %
We use this overlap definition and perform standard matching between predictions and groundtruth, with a minimum overlap threshold of $\alpha_{IoU}$.

When evaluating accuracy across all people, matched groundtruth boxes are true positives (TP), all unmatched groundtruth are false negatives (FNs, or misses), and unmatched detections are false positives (FP).
When evaluating accuracy on occluded people, only matched \textit{occluded} groundtruth boxes count as TPs, only unmatched \textit{occluded} groundtruth boxes count as FNs, and all unmatched detections count as FPs.
Intuitively, when evaluating metrics for occluded people, we do not penalize a detector for correctly detecting a visible person, but we \textit{do} penalize it for false positives that do not match any visible or occluded person.

We now describe how the $k$-vector of predictions is obtained: in addition to a state mean (first sample), our probabilistic method maintains covariances for $x$ and $z$ state variables which result in a 2D gaussian. Since these gaussians may extend incorrectly into freespace, we perform rejection sampling to accumulate $k$-1 predictions which respect freespace constraints. This gives us $P$. For baseline methods that are not probabilistic or do not have access to a depth map, we artificially simulate this distribution by tuning two scale factors that control the size of gaussians as a function of a bounding box's height. 
We tune these scale factors on MOT-17 train and use them throughout experiments. \tarasha{do we want to add why not y covariance?}

\textbf{Top-1 F1:} When $k=1$, this metric is simply the standard F1 metric. We additionally report this Top-1 F1 for occluded and \textit{all} people. We do not use the standard `average precision' (AP) metric as most detectors and trackers on the MOT and PANDA datasets do not report confidences.

{\bf IDF1:} To evaluate tracking, we report the standard IDF1 metric %
and also modify it for evaluating occluded people. Specifically, we divide the groundtruth tracks into visible and occluded segments, and perform matching only on the occluded segments. Once the tracks are matched, we compute IDTP as the number of matched occluded boxes, IDFP as the number of unmatched occluded \textit{or} visible predictions, and IDFN as the number of unmatched occluded groundtruth boxes. We similarly modify MOTA in the appendix.

To guide evaluation, we conduct a human vision experiment with 10 in-house annotators who annotated 991 boxes in 59 tracks with occlusion phases. Figure~\ref{fig:human_distribution} shows that annotators have lower consistency when labeling occluded people than visible people. To address this ambiguity in localizing occluded people, we choose a low $\alpha_{IoU}=0.5$ and $k=5$ in our experiments. 

}

\textbf{Implementation details.}
We empirically set parameters in our approach on MOT-17 train with Faster R-CNN \cite{ren2015faster} detections. The optimal thresholds for filtering forecasts on the train set are
$\alpha_{\textrm{delete}} = 0.88, \alpha_{\textrm{supp}} = 1.06$\footnote{Note that $\alpha_{\textrm{supp}} > 1$ allows the forecasted depth to be closer to the camera than the observed depth, accounting for potential noise in the depth estimator to reduce the number of forecasts that are suppressed.}. During occlusion we treat a person as a point, freezing its aspect ratio and height.
We fix $N_{age}$ to $30$.
The appendix presents further details of our method, \change{parameters and their tuning protocol}, including improvements by tuning $N_{age}$.
We tune on MOT-17 train and apply these tuned parameters on MOT-17 test, MOT-20, and PANDA.
\change{We find that our method and its hyperparameters tuned on the train set generalize well to the test set}.
\change{We use \cite{li2018megadepth} for monocular depth estimates, which has been shown to work well in the wild.
While these estimates can be noisy, 
we qualitatively find that the \textit{relative} depth orderings used in our approach
are fairly robust.}

\subsection{Oracle Study}
\label{sec:oracle}

\textbf{What is the impact of \textit{visible} detection on occluded detection?} 
We first evaluate an offline approach which uses groundtruth detections and tracks for visible people to (linearly) interpolate detections for occluded people in Table \ref{tab:oracle_simple}.
As this method perfectly localizes visible people, and most people in this benchmark are visible, it achieves a high overall Top-5 F1 of 98.0 (\Cref{tab:oracle_simple}, row 1). Additionally, despite using simple linear interpolation, this oracle also achieves a high Top-5 F1 of 87.3 for \textit{invisible} people.
This result indicates that although long-term forecasting of pedestrian trajectories may require higher-level reasoning \cite{scovanner2009learning,leal2011everybody,ma2017forecasting}, short-term occlusions may be modeled with simple linear models.

Next, we evaluate the same approach with detections from a Faster R-CNN \cite{ren2015faster} model in place of groundtruth (\Cref{tab:oracle_simple}, row 2).
This leads to a significant drop in both overall and occluded accuracy, indicating that improvements in \textit{visible} person detection can improve detection for invisible people.
Finally, although Occluded Top-5 F1 drops, it is significantly above chance, suggesting that current detectors equipped with appropriate trackers can detect invisible people.

\textbf{What is the impact of \textit{tracking} on occluded detection?}
So far, we have assumed oracle linking of detections, allowing for linear interpolation of bounding boxes to detect people through occlusion.
We now evaluate the impact of using an online tracker, equipped with re-identification, on detecting occluded people.
Removing the oracle results in a drastic drop in accuracy: the Top-5 F1 score for occluded people drops by over 30 points (87.3 to 53.3, \Cref{tab:oracle_simple} row 3) using groundtruth detections, and 14 points with Faster R-CNN detections (46.4 to 32.2, \Cref{tab:oracle_simple} row 4).
Despite this significant drop in Occluded Top-5 F1, the overall Top-5 F1 is significantly more stable (from 98.0 to 92.3 for groundtruth detections and 70.5 to 69.9 for Faster R-CNN), showing that \textit{overall} person detection and tracking underemphasizes the importance of detecting occluded people.

\textbf{Can online approaches work?}
These results indicate that in the offline setting, existing visible-person detection and tracking approaches are can detecting invisible people via interpolation.
We now evaluate a simple \textit{online} approach, which uses an off-the-shelf visible person detector (Faster R-CNN), equipped with a tracker (DeepSORT) and linear (constant velocity) forecasting for detecting invisible people (\Cref{tab:oracle_simple}, row 5).
Moving to an online setting results in a similar Top-5 F1 score but significantly reduces the precision for occluded persons, from 60.8 to 29.5. This is expected as even though linear forecasting recalls slightly more number of boxes than offline interpolation (recall from 21.9 to 30.2), its naive nature results in many more false positives resulting in a much lower precision and therefore, a similar F1 score. In Section \ref{sec:ablation}, we present simple modifications to this approach that recover much of this performance gap.

\renewcommand{\tabcolsep}{0.1cm}
\begin{table}[t]
\caption{Detection and tracking results on MOT-17 \cite{milan2016mot16}, MOT-20 \cite{dendorfer2020mot20} and PANDA \cite{wang2020panda} train. We evaluate on public detections provided with MOT-17 (DPM \cite{felzenszwalb2009object}, FRCNN \cite{ren2015faster}, SDP \cite{yang2016exploit}), two trackers that operate on public detections (Tracktor++ \cite{bergmann2019tracking}, MIFT \cite{huang2020refinements}), and CenterTrack \cite{zhou2020tracking} which does not use public detections. We use (public FRCNN, \textit{visible} groundtruth) detections for (MOT-20, PANDA).
Our approach improves on occluded people across all trackers. }
\centering
\resizebox{\linewidth}{!}{
\begin{tabular}
{clllcccc}
\toprule[1.5pt]
& & \multicolumn{2}{c}{Top-5 F1} & \multicolumn{2}{c}{Top-1 F1} & \multicolumn{2}{c}{IDF1}\\

\cmidrule(lr){3-4}
\cmidrule(lr){5-6}
\cmidrule(lr){7-8}
& & Occl & All & Occl & All & Occl & All\\
\midrule[.5pt]
 \parbox[t]{3mm}{\multirow{12}{*}{\rotatebox[origin=c]{90}{MOT-17}}} & DPM & 17.2  & 46.7  & 13.2 & 46.5 & 2.9 & 36.9\\
&  + Ours &24.6 {\scriptsize(+7.4)} & 49.3 {\scriptsize(+2.6)} & 17.4 & 48.4 & 7.2 & 36.8\\

& FRCNN & 28.4  & 68.5  & 20.1 & 67.4 & 1.5 & 55.6\\
&  + Ours &39.8 {\scriptsize(+11.4)} & 70.5 {\scriptsize(+2.0)} & 26.7 & 68.5 & 10.5 & 54.8\\

& SDP & 45.2  & 80.5  & 35.8 & 79.8 & 10.9 & 64.6\\
&  + Ours &51.2 {\scriptsize(+6.0)} & 80.8 {\scriptsize(+0.3)} & 38.5 & 79.4 & 17.0 & 64.7\\

& Tracktor++ & 32.4  & 77.0  & 22.7 & 76.8 & 1.3 & 65.1\\
&  + Ours & 45.4 {\scriptsize(+13.0)} & 77.2 {\scriptsize(+0.2)} & 33.2 & 76.5 & 15.6 & 66.8\\

& MIFT & 37.8  & 75.9  & 29.9 & 75.1 & 9.4 & 61.7\\
&  + Ours &44.9 {\scriptsize(+7.1)} & 75.6 {\scriptsize(-0.3)} & 33.8 & 74.3 & 16.5 & 62.6\\

& CTrack & 38.7  & 84.8  & 29.4 & 84.2 & 5.4 & 65.0\\
&  + Ours &47.9 {\scriptsize(+9.2)} & 84.4 {\scriptsize(-0.4)} & 36.4 & 83.4 & 16.2 & 70.2\\
\midrule[0.5pt]

\parbox[t]{3mm}{\multirow{2}{*}{\rotatebox[origin=c]{90}{\scriptsize MOT-20}}}
& FRCNN  & 42.5  & 71.2  & 27.5 & 70.7 & 2.9 & 42.2\\
&  + Ours & 46.1 {\scriptsize(+3.6)} & 71.5 {\scriptsize(+0.3)} & 28.6 & 70.9 & 5.0 & 42.0\\
\midrule[0.5pt] 

\parbox[t]{3mm}{\multirow{2}{*}{\rotatebox[origin=c]{90}{\scriptsize PANDA}}} 
& GT (visible) & 45.5  & 90.6  & 30.5 & 90.5 & 2.5 & 70.2\\
&  + Ours& 49.5 {\scriptsize(+4.0)} & 90.5 {\scriptsize(-0.1)} & 34.1 & 90.3 & 4.6 & 62.1\\

\bottomrule[1.5pt]
\end{tabular}}
\label{tb:sota}
\end{table}

\subsection{Comparison to Prior Work}
\label{sec:sota}
\newlength{\oldintextsep}
\setlength{\oldintextsep}{\intextsep}
\setlength{\intextsep}{4ex}%
Next, we apply our approach to the output of existing methods to evaluate its improvement over prior work.
Table \ref{tb:sota} shows results on the MOT-17 train set, showing our approach improves significantly in Occluded Top-5 F1 ranging from 6.0 to 13.0 points, while maintaining the overall F1.
\newstuff{Detecting invisible people requires reliable amodal detectors for visible people (ref. Section \ref{sec:oracle}). 
For this reason, we use \textit{visible} groundtruth detections from PANDA, similar to the oracle experiments in~\Cref{sec:oracle}, as no public set of amodal detections come with PANDA (unlike MOT-17 or MOT-20). }%
\tarasha{change numbers in this paragraph after Table 4 is complete}%
\newstuff{Table \ref{tb:sota} shows that our method improves the detection of occluded people by 4.0\% on PANDA using groundtruth visible detections and by 3.6\% on MOT-20 using the Faster-RCNN public detections. We explicitly do not tune our hyperparameters 
for these two datasets, showing that our method is robust to changes in video data distribution. 
MOT-20 and PANDA contain a few sequences with top-down views, where occlusions are rare. We disable our depth and occlusion reasoning on such sequences; please see appendix for details.
}

\renewcommand{\tabcolsep}{0.1cm}
\begin{table}[t]
\centering
\resizebox{0.9\linewidth}{!}{
    \begin{tabular}{clcccccc}
\toprule[1.5pt]
 & & \multicolumn{2}{c}{Top-5 F1} & \multicolumn{2}{c}{Top-1 F1} & \multicolumn{2}{c}{IDF1} \\
\cmidrule(lr){3-4}
\cmidrule(lr){5-6}
\cmidrule(lr){7-8}
 & & Occl & All & Occl & All & Occl & All\\
\midrule[.5pt]
        \parbox[t]{3mm}{\multirow{6}{*}{\rotatebox[origin=c]{90}{MOT-17}}} & Ours & {\color{blue} 43.4} & {\color{darkgreen} 76.8} & {\color{blue} 31.4} & {\color{darkgreen} 75.6} & {\color{blue} 14.7} & {\color{red}58.7}\\
        & MIFT \cite{huang2020refinements} & {\color{red}38.4} & {\color{red}77.3} & {\color{red}29.7} & {\color{red}76.7} & {\color{red}10.4} & 56.4\\
        & UnsupTrack \cite{karthik2020simple} & {\color{darkgreen} 35.9} & {\color{blue} 78.1} & {\color{darkgreen} 26.6} & {\color{blue} 77.4} & {\color{darkgreen} 9.7} & {\color{blue} 62.6}\\
        & GNNMatch \cite{papakis2020gcnnmatch} & 35.2 & 74.3 & 26.3 & 73.7 & 6.9  & 56.1\\
        & GSM\_Tracktor \cite{ijcai2020-74} & 35.4 & 73.8 & 26.2 & 73.2 & 7.4  & {\color{darkgreen} 57.8}\\
        & Tracktor++ \cite{bergmann2019tracking} & 33.3 & 73.3 & 24.8 & 73.0 & 5.2 & 55.1\\
    \midrule[.5pt]
        \parbox[t]{3mm}{\multirow{4}{*}{\rotatebox[origin=c]{90}{MOT-20}}} & Ours & {\color{blue} 46.9} & {\color{blue} 76.7} & {\color{red}33.3} & {\color{red}75.2} & {\color{blue} 11.2} & {\color{blue} 51.1}\\
        & Tracktor++ \cite{bergmann2019tracking} & {\color{red}44.2} & {\color{red}76.0} & {\color{blue} 34.2} & {\color{blue} 75.3} & {\color{red}10.2} & {\color{darkgreen} 48.8}\\
        & UnsupTrack \cite{karthik2020simple} & {\color{darkgreen} 41.7} & {\color{darkgreen} 71.4} & {\color{darkgreen} 30.9} & {\color{darkgreen} 70.8} & {\color{darkgreen} 9.6} & {\color{red}50.6}\\
        & SORT20  \cite{wojke2017simple} & 38.5 & 65.2 & 27.3 & 63.6 & 8.8 & 45.1\\
    \bottomrule[1.5pt]
    \end{tabular}}
\caption{Results on MOT-17 and MOT-20 test set. The {\color{blue} best}, {\color{red} second-best} and {\color{darkgreen} third-best} methods are highlighted.
}\label{tb:sota_test}
\end{table}
As \newstuff{MOT-17 and MOT-20 test labels are held out, we worked with the MOTChallenge authors to implement our metrics on the test server. 
Table \ref{tb:sota_test} shows that MIFT\footnote{MIFT is referred to as ISE\_MOT17R on the MOT-17 and MOT-20 leaderboards}\cite{huang2020refinements} and Tracktor++ \cite{bergmann2019tracking} achieve the highest Occluded Top-5 F1 amongst prior online approaches on MOT-17 and MOT-20 test respectively.
Applying our approach on top of these methods improves results significantly by 5.0\% to 43.4 F1 and by 2.7\% to 46.9 F1, leading to a new state-of-the-art for occluded person detection on MOT-17 and MOT-20 test. }

\begin{figure*}[t]
  \centering
  \includegraphics[width=0.93\linewidth]{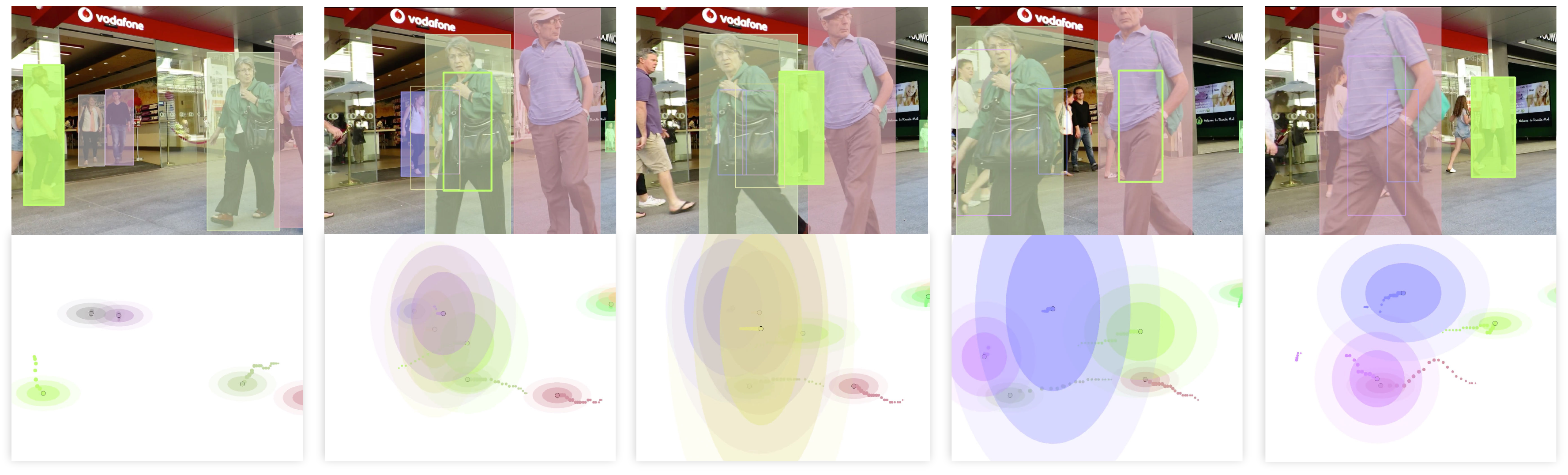}
  \caption{
  Our probabilistic model reports a {\em distribution} over 3D location during occlusions. We visualize (occluded, visible) detection with (outlined, filled-in) bounding boxes ({\bf top}). We provide ``birds-eye-view" top-down visualizations of Gaussian distributions over 3D object centroids with covariance ellipses ({\bf bottom}). During occlusion, variance grows roughly linearly with the number of consecutively-occluded frames. We are also able to correctly predict depth of occluded people in the top down view, e.g. in the second last frame, which would not be possible with single-frame monocular depth estimates. \newstuff{During evaluation, we truncate the uncertainty using our freespace estimates (not visualized).}
  }
  \label{fig:result}
\end{figure*}

\change{\Cref{tb:sota} shows that our method consistently improves occluded F1. However, it sometimes results in a drop in overall accuracy. We attribute this to the increased number of false positives introduced while tackling the challenging task of detecting invisible people.
These false positives for invisible people are counted as false positives for \textit{all} people, whether visible or invisible.
This causes
existing metrics to penalize methods for even \textit{trying} to detect invisible people.
In safety critical applications, where worst-case accuracy may be more appropriate, our approach significantly improves during complete occlusions by up to 11.4\% on MOT-17, while mildly decreasing average accuracy by 0.5\%.
}

\subsection{Ablation Study}
\label{sec:ablation}
We now study the impact of each component of our approach in Table \ref{tb:ablation_simple}, focusing on the Occluded Top-5 F1 metric using Faster R-CNN detections on the MOT-17 train set. 
First, we show that the DeepSORT tracker, upon which our approach is built, results in a 28.4 Occluded Top-5 F1.
Reporting the internal, linear forecasts from the tracker increases the score to 29.8, driven primarily by a 12.5\% improvement in recall.
Compensating for camera motion provides another 2.4\% improvement.
Next, leveraging depth cues to incorporate freespace constraints, as detailed in Section \ref{sec:temporalnoise}, improves accuracy by 3.5\%, driven primarily by a 14.6\% jump in precision, indicating that this component drastically reduces false positives.
Finally, we add depth-aware process noise to handle perspective transformations between 2D and 3D coordinates, which leads to an improvement of 4.1\%, resulting in a final score of 39.8. \newstuff{Only a 1.0\% improvement in F1 as compared to 4.1\% with Top-5 F1 suggests that our uncertainty estimates are significantly improved by the depth-aware process noise scaling.} In all, our approach leads to an improvement of 11.4\% over the baseline. Figure \ref{fig:result} presents a sample result from our approach, where the person in the green bounding box is detected throughout two full occlusion phases, marked with an unfilled box.

\newstuff{One concern with our approach might be that the average depth inside a person's bounding box may contain pixels from the background or an occluder. To verify the impact of this, we evaluate a variant where we use segmentation masks for all the bounding boxes in MOT-17's FRCNN public detections using MaskRCNN \cite{he2017mask}. We initialize the $z$ state variable in the model with the average depth inside this mask. On doing so, the Top-1 occluded F1 increases from 26.7 to 27.3, indicating that masks can help with estimating the person's depth, but boxes are a reasonable approximation.}
We kindly refer the reader to the appendix for further ablative analysis, including an analysis of more recent depth estimators, ablations on moving \vs stationary sequences, and failure cases.

\renewcommand{\tabcolsep}{0.1cm}
\begin{table}[t]
\caption{MOT-17 train ablations. Each row adds a component to the row above. `Dep. noise' is depth-aware noise.\vspace{-1em}}
    \label{tb:ablation_simple}
\centering
\resizebox{\linewidth}{!}{
    \begin{tabular}{lcccccccc}
\toprule[1.5pt]
 & \multicolumn{4}{c}{Top-5} & \multicolumn{2}{c}{Top-1 F1} & \multicolumn{2}{c}{IDF1} \\
  \cmidrule(lr){2-5}
\cmidrule(lr){6-7}
\cmidrule(lr){8-9}

 & Occl F1 & Occl Prec & Occl Rec & All F1 & Occl & All & Occl & All\\

\midrule[.5pt]
    DeepSORT & 28.4 {\scriptsize$\pm$0.1} & 
    71.9 {\scriptsize$\pm$0.2} & 17.7 {\scriptsize$\pm$0.1} & 68.5 {\scriptsize$\pm$0.0} & 20.1 & 67.4 & 1.5 & 55.6\\
    + Forecast & 29.8 {\scriptsize$\pm$0.2}  & 
    29.5 {\scriptsize$\pm$0.4} & 30.2 {\scriptsize$\pm$0.1}& 69.4 {\scriptsize$\pm$0.0} & 20.9 & 66.5 & 7.6 & 53.3\\
    + Egomotion & 32.2 {\scriptsize$\pm$0.2} & 
    33.1 {\scriptsize$\pm$0.3} & 31.3 {\scriptsize$\pm$0.1} & 70.4 {\scriptsize$\pm$0.0} & 23.2 & 67.9 & 9.1 & 54.5\\
    + Freespace & 35.7 {\scriptsize$\pm$0.0} & 
    47.7 {\scriptsize$\pm$0.1} & 28.6 {\scriptsize$\pm$0.0} & 70.4 {\scriptsize$\pm$0.0} & 25.7 & 68.4 & 9.7 & 55.0\\
    + Dep. noise & 39.8 {\scriptsize$\pm$0.2} & 
    52.6 {\scriptsize$\pm$0.6} & 32.0 {\scriptsize$\pm$0.0} & 70.5 {\scriptsize$\pm$0.1} & 26.7 & 68.5 & 10.5 & 54.8 \\
    \bottomrule[1.5pt]
    \end{tabular}}
\end{table}

\textbf{Forecasting:} 
We evaluate replacing our linear forecaster with state-of-the-art forecasters.
We supply these forecasters with a birds-eye-view representation of visible person trajectories.
As these forecasters forecast only the birds-eye-view ($x, z$) coordinates, we rely on our approach's estimates of the height, width, and $y$ coordinate.
We evaluate two trajectory forecasting approaches for crowded scenes, Social GAN (SGAN) \cite{gupta2018social} and STGAT \cite{huang2019stgat}. 
SGAN and STGAT result in Occluded Top-5 F1 scores of 36.0 and 36.4 respectively. While this improves over the baseline at 28.4, it underperforms our linear forecaster at 39.8.
This suggests that simple linear models suffice for short, frequent occlusions. We refer the reader to the appendix for more details and analysis.

\section{Discussion}
We propose the task of detecting fully-occluded objects from monocular cameras in an online manner.
Our experiments show that current detection and tracking approaches struggle to find occluded people, dropping in accuracy from 68\% to 28\% F1. Our oracle experiments reveal that interpolating across tracklets in an offline setting noticeably improves F1, but the task remains difficult because underlying object detectors do not perform well during large occlusions. We propose an online approach that forecasts the trajectories of occluded people, exploiting depth estimates from a monocular depth estimator to better reason about potential occlusions. Our approach can be applied to the output of existing detectors and trackers, leading to significant accuracy gains of 11\% over the baseline, and 5\% over state-of-the-art. We hope our problem definition and initial exploration of this safety-critical task encourages others to do so as well.

\section*{Acknowledgements}

We thank Gengshan Yang for his help with generating 3D visuals, Patrick Dendorfer for incorporating our metrics with the MOT challenge server, and Xueyang Wang for sharing the low-resolution version of the PANDA dataset. We thank Laura Leal-Taix\'e and Simon Lucey for insightful discussions, internal reviewers at the Robotics Institute, CMU for reviewing early drafts, and participants of the human vision experiment. This work was supported by the CMU Argo AI Center for Autonomous Vehicle Research, the Defense Advanced Research Projects Agency (DARPA) under Contract No. HR001117C0051, and the National Science Foundation (NSF) under grant number IIS-1618903.

\appendix

\begin{center}
    \Large\bfseries Appendix
\end{center}

We now provide further analysis of our method and implementation details of our experiments.
\Cref{sec:supp_ablation} presents additional ablation studies of our method.
\Cref{sec:supp_mota} extends our tracking evaluation to use the popular MOTA (multi-object tracking accuracy) metric.
\Cref{sec:supp_human_vision} provides details regarding our human vision experiment, which analyzes people's ability to detect and localize highly occluded objects.
\Cref{sec:supp_panda_mot20} discusses the experimental setup for PANDA and MOT-20 datasets, and \Cref{sec:supp_pseudocode} presents pseudocode of our final depth-aware tracking algorithm.

\section{Ablation Study}
\label{sec:supp_ablation}
In this section, we analyze the impact of using different depth estimators (\Cref{sec:supp_ablation_depth}), segmentation masks in place of bounding boxes for estimating average depth (\Cref{sec:supp_ablation_masks}), more sophisticated forecasters (\Cref{sec:supp_ablation_forecasting}), the performance of our method on moving \vs stationary cameras (\Cref{sec:supp_ablation_moving}), and, finally, the importance of different hyperparameters (\Cref{sec:supp_ablation_params}).

\subsection{Monocular Depth Estimators}
\label{sec:supp_ablation_depth}

\newstuff{Our method relies on an off-the-shelf monocular depth estimator to enable occlusion reasoning in 3D.
In our main paper, we used the MegaDepth~\cite{li2018megadepth} estimator throughout our experiments.
Here, we evaluate whether recent advances in monocular depth estimation provide more reliable \textit{relative} depth estimates of people as used by our method.
Specifically, we replace the MegaDepth estimator with the MannequinChallenge~\cite{li2019learning} and MIDAS~\cite{lasinger2019towards} depth estimators in our method.
We evaluate on MOT-17 using the Faster-RCNN set of public detections, and set all hyperparameters in our pipeline to their default values and disable the depth-aware noise scaling. This simple variant of our pipeline allows us to evaluate the quality of depth estimates from each of the three methods.
Table \ref{tab:mono} shows that the per frame depth estimator from Mannequin Challenge~\cite{li2019learning} does worse than MegaDepth~\cite{li2018megadepth} by 1.2 Top-5 F1 for invisible people and MIDAS~\cite{lasinger2019towards} similarly does worse by 1.0 point. By the standard Top-1 F1 metric, these estimators degrade accuracy by 1.2 and 0.2 points respectively.
As this simple variant of our pipeline is aimed at evaluating the relative depth orderings output from the depth estimators, these results suggest that while these depth estimators have become more accurate and generalizable over the years, the relative depth orderings of objects has not significantly improved.

Since monocular depth estimators can take as input images of varying sizes, we evaluate the effect of using higher resolution images as input to the estimator.
Using a higher resolution input can increase the size of smaller objects in the scene (e.g., people far away), potentially allowing depth estimators to output more precise depth estimates.
We evaluate using higher resolutions as input with the MIDAS~\cite{lasinger2019towards} estimator in \Cref{tab:monores}.
By default, we resize images to a resolution of 512$\times$384 pixels (`1x', the resolution MIDAS is trained with) from their original resolution of 1920$\times$1080.
We evaluate MIDAS \cite{lasinger2019towards} at $2 \times$ and $3 \times$ this default resolution and find in that doing so improves the Top-5 F1 for invisible peopleby 3.1\%.
We note here that this is not the case with the other two depth estimators~\cite{li2018megadepth,li2019learning} whose performance decreases or stagnates with higher resolutions (not shown).}

\begin{table}
\renewcommand{\tabcolsep}{0.1cm}
    \centering
\resizebox{\linewidth}{!}{\begin{tabular}{lcccccc}
    \toprule[1.5pt]
    & \multicolumn{2}{c}{Top-5 F1} & \multicolumn{2}{c}{Top-1 F1} & \multicolumn{2}{c}{IDF1} \\
     \cmidrule(lr){2-3}
\cmidrule(lr){4-5}
\cmidrule(lr){6-7}
    Depth est. & Occl & All & Occl & All & Occl & All \\
    \midrule[.5pt]
    MegaDepth \cite{li2018megadepth} &
          35.4{\scriptsize$\pm$0.2} & 69.8{\scriptsize$\pm$0.0} & 26.7 & 68.4 & 9.5 & 53.3 \\
    Mannequin \cite{li2019learning}  &
         34.2{\scriptsize$\pm$0.2} & 69.4{\scriptsize$\pm$0.0} & 25.5 & 68.0 & 8.5 & 53.3 \\
    MIDAS \cite{lasinger2019towards} &
          34.4{\scriptsize$\pm$0.1} & 69.5{\scriptsize$\pm$0.0} & 26.5 & 68.2  & 9.1 & 53.8 \\
    \bottomrule[1.5pt]
    \end{tabular}}
    \caption{Comparison of different monocular depth estimators used in our pipeline. More recent depth estimators do not seem to provide more reliable \textit{relative} depth orderings, which are used by our method.}
    \label{tab:mono}
\end{table}

\begin{table}
\renewcommand{\tabcolsep}{0.1cm}
    \begin{tabular}{lccccccc}
    \toprule[1.5pt]
    & & \multicolumn{2}{c}{Top-5 F1} & \multicolumn{2}{c}{Top-1 F1} & \multicolumn{2}{c}{IDF1} \\
     \cmidrule(lr){3-4}
\cmidrule(lr){5-6}
\cmidrule(lr){7-8}

    Depth & Res. & Occl & All & Occl & All & Occl & All \\
    \midrule[.5pt]
    MIDAS & 1x &
          34.4{\scriptsize$\pm$0.1} & 69.5{\scriptsize$\pm$0.0} & 26.5 & 68.2  & 9.1 & 53.8 \\
    MIDAS & 2x &
          35.5{\scriptsize$\pm$0.2} & 70.0{\scriptsize$\pm$0.0} & 27.0 & 68.5 & 9.8 & 53.9 \\
    MIDAS & 3x &
          37.5{\scriptsize$\pm$0.2} & 69.9{\scriptsize$\pm$0.0} & 27.0 & 68.2 & 10.8 & 53.9 \\
    \bottomrule[1.5pt]
    \end{tabular}
    \caption{We evaluate a recent depth estimator, MIDAS \cite{lasinger2019towards}, at varying input resolutions. At higher resolutions (3x), the estimator improves Top-5 F1 by 3.1 points, suggesting higher resolutions can improve depth estimates, likely by providing more reliable relative depths for faraway pedestrians.}
    \label{tab:monores}
\end{table}

\subsection{Boxes vs Masks}
\label{sec:supp_ablation_masks}
\newstuff{
Our method estimates a person's depth by taking the average of the depth estimates within the person's bounding box.
However, these pixels may contain background regions, leading to incorrect depth estimates.
To address this, we evaluate a variant which uses an off-the-shelf instance segmentation method to only compute the average depth within a predicted person mask.
To do this, we pass the Faster R-CNN public detections from MOT-17 as proposals into the mask head of Mask R-CNN~\cite{he2017mask}.
Occasionally, this instance segmentation method may fail to produce a reasonable mask for a person.
We design a simple strategy for detecting a common failure case:
if the output segmentation mask covers less than 25\% of the bounding box (in cases where the people are too small or out-of-distribution), we discard the predicted mask and treat the full bounding box as the mask.
We do not use masks for the forecasted boxes of occluded people, as these boxes cover unknown occluders.
In Table \ref{tab:boxmasks}, we find that masks modestly help our method, increasing Top-5 and Top-1 F1 by 0.6 and 0.8 points for occluded people.
Interestingly, we also see an increase in overall F1 by the same amount.
}
\begin{table}
\renewcommand{\tabcolsep}{0.1cm}
    \begin{tabular}{lcccccc}
    \toprule[1.5pt]
    & \multicolumn{2}{c}{Top-5 F1} & \multicolumn{2}{c}{Top-1 F1} & \multicolumn{2}{c}{IDF1} \\
     \cmidrule(lr){2-3}
\cmidrule(lr){4-5}
\cmidrule(lr){6-7}

    & Occl & All & Occl & All & Occl & All \\
    \midrule[.5pt]
    Boxes & 39.8 {\scriptsize$\pm$0.2} & 70.5 {\scriptsize$\pm$0.1} & 26.7 & 68.5 & 10.5 & 54.8 \\
    Masks &
          40.6{\scriptsize$\pm$0.3} & 71.3{\scriptsize$\pm$0.0} & 27.3 & 69.1 & 11.0 & 54.7\\
    \bottomrule[1.5pt]
    \end{tabular}
    \caption{Replacing boxes by masks for getting mean depth of a person only helps by a small amount suggesting that boxes can reasonably replace masks.}
    \label{tab:boxmasks}
\end{table}

\subsection{Forecasting Approaches}
\label{sec:supp_ablation_forecasting}
As described in the main paper, we use a constant velocity forecaster in our probabilistic approach. In Sec 4.3, we showed that replacing our our simple linear forecaster with more sophisticated state-of-the-art forecasters that exploit social cues did not improve performance.
Here, we provide implementation details for these experiments, and analyze different variants.
The approaches discussed in the main paper,  SGAN \cite{gupta2018social} and STGAT \cite{huang2019stgat} are supplied the top-down views from our algorithm. Both SGAN and STGAT forecast 20 samples and then choose the closest trajectory to the groundtruth from these 20.
This advantage is not feasible for an online approach where groundtruth cannot be supplied to the algorithm.
To simulate the online setting, we sample the mean trajectory from these approaches by requesting the trajectory corresponding to the zero noise vector.
We calculate an approximate average scale factor of 20.0 between the trajectory values learnt by these models and the trajectory values available for input from our method, which we use to scale down our input values.
Additionally, each of these methods has an 8- and 12-timestep forecasting model.
In the main paper, we report the best of these models for both approaches and report other models in Table~\ref{tab:forecasting}.
For STGAT, the 8- and 12-timestep models used are trained on the ETH~\cite{pellegrini2009you} dataset and for SGAN, the 8- and 12-timestep models are trained on the ZARA1~\cite{lerner2007crowds} dataset.
Each of these models is made to predict for 30-timesteps by supplying the last 8 forecasted timesteps iteratively.
The occlusion phase may not last 30 timesteps for all people.
We therefore use the information from our pipeline about the number of occluded timesteps and replace the x and z values from the output of our pipeline with SGAN and STGAT's forecasted x and z values.
In Table \ref{tab:forecasting}, we additionally report the performance of the methods when we provide past trajectories of \textit{multiple} people as input, allowing the method to leverage social cues.
For the Top-5 evaluation, we use the blind baseline described in Sec. 4 of our main paper.
The conclusion remains that simple linear models suffice for short, frequent occlusions as our approach always performs better than any of the social forecasting settings of SGAN and STGAT.
 
\begin{table}
\renewcommand{\tabcolsep}{0.1cm}
    \begin{tabular}{clcccccc}
    \toprule[1.5pt]
    & & \multicolumn{2}{c}{Top-5 F1} & \multicolumn{2}{c}{Top-1 F1} & \multicolumn{2}{c}{IDF1} \\
     \cmidrule(lr){3-4}
\cmidrule(lr){5-6}
\cmidrule(lr){7-8}

    & & Occl & All & Occl & All & Occl & All \\
    \midrule[.5pt]
     \parbox[t]{3mm}{\multirow{4}{*}{\rotatebox[origin=c]{90}{Single}}} & SGAN-8 &
          35.4{\scriptsize$\pm$0.2} & 70.2{\scriptsize$\pm$0.0} & 24.6 & 67.8 & 8.9 & 54.3 \\
    & SGAN-12 &
           35.0{\scriptsize$\pm$0.1} & 70.1{\scriptsize$\pm$0.0} & 24.2 & 67.7 & 8.7 & 54.2 \\
    & STGAT-8 &
           35.1{\scriptsize$\pm$0.1} & 70.1{\scriptsize$\pm$0.0} & 24.5 & 67.6 & 8.6 & 54.3 \\
    & STGAT-12 &
           35.6{\scriptsize$\pm$0.2} & 70.3{\scriptsize$\pm$0.0} & 24.7 & 67.9 & 9.1 & 54.4 \\
         \midrule[0.5pt]
     \parbox[t]{3mm}{\multirow{4}{*}{\rotatebox[origin=c]{90}{Multi}}} & SGAN-8 &
           36.0{\scriptsize$\pm$0.2} & 70.3{\scriptsize$\pm$0.0} & 24.8 & 67.9 & 9.2 & 54.4 \\
    & SGAN-12 &
           36.0{\scriptsize$\pm$0.3} & 70.3{\scriptsize$\pm$0.0} & 24.9 & 67.9 & 9.3 & 54.4 \\
    & STGAT-8 &
           36.2{\scriptsize$\pm$0.3} & 70.3{\scriptsize$\pm$0.0} & 24.5 & 67.8 & 8.8 & 54.3 \\
    & STGAT-12 &
           36.4{\scriptsize$\pm$0.1} & 70.4{\scriptsize$\pm$0.0} & 24.8 & 67.9 & 9.2 & 54.4 \\
    \bottomrule[1.5pt]
    \end{tabular}
    \caption{MOT-17 train forecasting ablations with state-of-the-art social forecasting models.}
    \label{tab:forecasting}
\end{table}

\subsection{Moving vs Stationary Camera Sequences}
\label{sec:supp_ablation_moving}
In the MOT-17 dataset, 3 camera sequences are stationary and 4 are captured from a moving camera. We separately study the effect of using different components of our pipeline on these sets of camera sequences. \Cref{tb:ablation_moving_stationary} shows that compensating for camera egomotion and filtering estimates lying in freespace helps the moving camera sequences by 4.5\% and 4.0\% Occluded Top-5 F1 respectively while for the stationary camera sequences, enforcing smoother tracks for faraway objects and filtering freespace estimates helps by 3.6\% and 2.0\% F1 respectively.

\renewcommand{\tabcolsep}{0.15cm}
\begin{table}
\centering
\resizebox{\linewidth}{!}{
    \begin{tabular}{lcccccccc}
\toprule[1.5pt]
 & \multicolumn{4}{c}{Top-5} & \multicolumn{2}{c}{Top-1 F1} & \multicolumn{2}{c}{IDF1} \\
\cmidrule(lr){2-5}
\cmidrule(lr){6-7}
\cmidrule(lr){8-9}
 & \makecell{Occl\\F1} & \makecell{Occl\\Prec} & \makecell{Occl\\Rec} & \makecell{All\\F1} & Occl & All & Occl & All\\
\midrule[.5pt]
    \multicolumn{9}{c}{Moving sequences} \\\midrule
    DeepSORT       & 27.3 {\scriptsize$\pm$0.3}   & 49.7 & 18.8 & 72.4 {\scriptsize$\pm$0.0}   & 17.3   & 67.0   & 2.2   & 56.5 \\                                                    + Forecast     & 21.3 {\scriptsize$\pm$0.1}   & 15.4 & 34.6 & 68.4 {\scriptsize$\pm$0.1}   & 13.3   & 63.6   & 5.6   & 50.2\\
    + Egomotion    & 25.8 {\scriptsize$\pm$0.0}   & 19.4 & 38.7 & 71.3 {\scriptsize$\pm$0.0}   & 17.1   & 66.9   & 8.7   & 53.2\\                                                     + Freespace    & 29.8 {\scriptsize$\pm$0.3}   & 28.0 & 31.8 & 72.8 {\scriptsize$\pm$0.0}   & 19.9   & 69.2   & 9.4   & 55.2\\
    + Dep. noise   & 34.3 {\scriptsize$\pm$0.1}   & 32.8 & 35.9 & 73.3 {\scriptsize$\pm$0.1}   & 20.2   & 69.4   & 9.8   & 55.9\\                                                     \midrule
    \multicolumn{9}{c}{Stationary sequences} \\\midrule
    DeepSORT       & 29.2 {\scriptsize$\pm$0.1}   & 94.0 & 17.3 & 66.2 {\scriptsize$\pm$0.0}   & 21.7   & 65.9   & 1.1    & 55.0 \\
    + Forecast     & 39.1 {\scriptsize$\pm$0.4}   & 62.2 & 28.5 & 70.2 {\scriptsize$\pm$0.0}   & 28.7   & 68.6   & 10.1   & 55.4\\
    + Egomotion    & 38.0 {\scriptsize$\pm$0.1}   & 60.2 & 27.8 & 69.8 {\scriptsize$\pm$0.0}   & 28.5   & 68.5   & 9.6    & 55.3\\
    + Freespace    & 40.0 {\scriptsize$\pm$0.0}   & 76.1 & 27.1 & 68.9 {\scriptsize$\pm$0.0}   & 30.3   & 67.9   & 10.0   & 54.9\\
    + Dep. noise   & 43.6 {\scriptsize$\pm$0.3}   & 78.7 & 30.2 & 68.8 {\scriptsize$\pm$0.0}   & 31.4   & 67.9   & 11.2   & 54.1\\
    \bottomrule[1.5pt]
    \end{tabular}}
    \caption{MOT-17 train ablations for moving \vs stationary camera sequences.}
    \label{tb:ablation_moving_stationary}
\end{table}

\subsection{Hyperparameter tuning}
\label{sec:supp_ablation_params}

\begin{figure}[t] %
  \centering
  \includegraphics[width=0.8\textwidth]{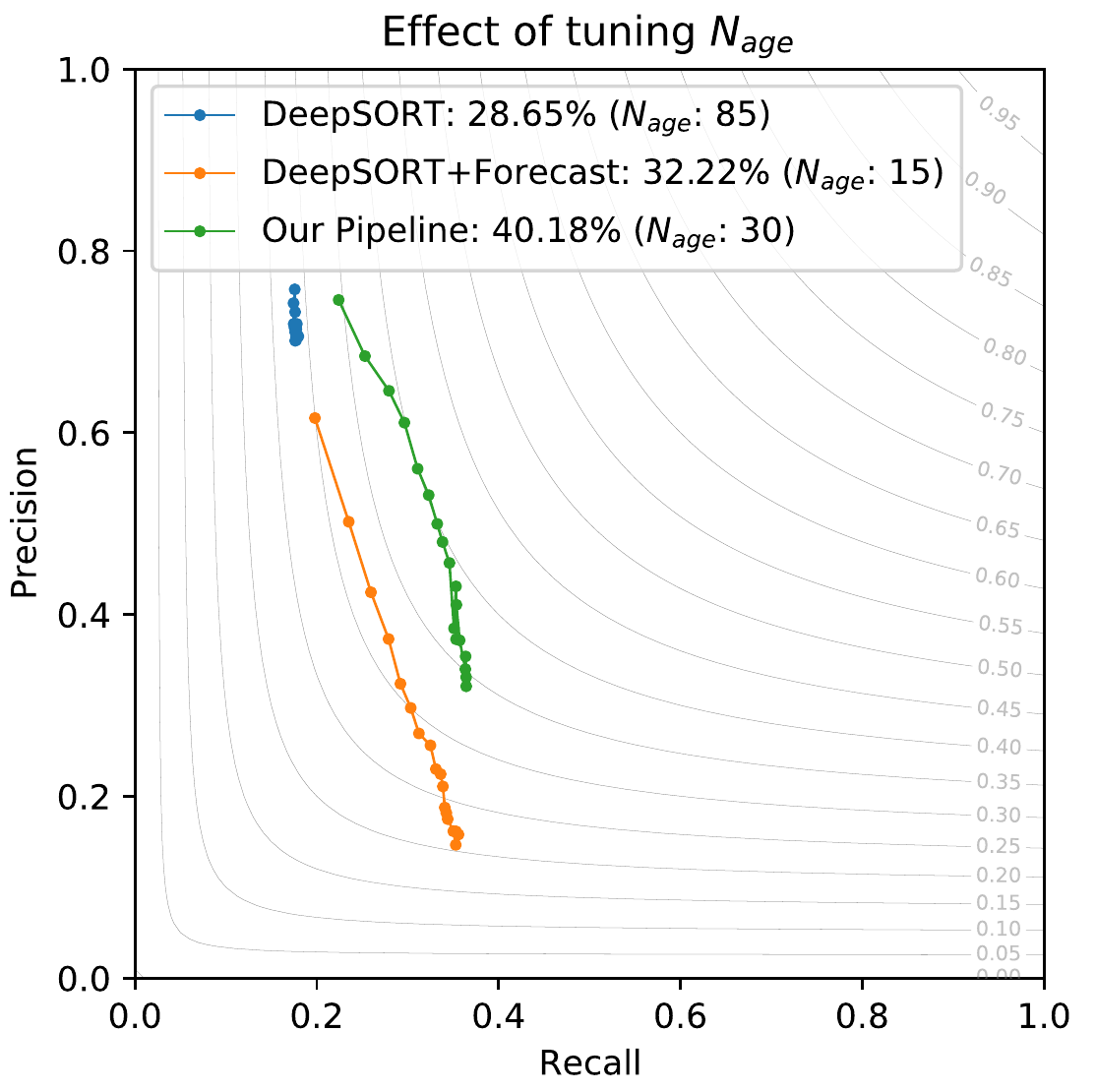}
  \caption{Detecting occluded people is sensitive to the threshold used to declare a detection-under-high-occlusion. We fix the number of $N_{age}$ frames that a track is allowed to be in an occluded state. By increasing $N_{age}$, we can tradeoff precision and recall in invisible-people-detection which results in a ``PR-curvelet''. The curvelets represent the experiments in rows 1, 2 and 5 of Table 4 in the main paper.
  }
  \label{fig:pr}
\end{figure}

 We describe a few parameters of our approach and how to tune them, in addition to the ones described in the paper. The $N_{age}$ parameter in our pipeline controls the number of frames that an occluded track is forecasted for before it is deleted.
We show in Figure~\ref{fig:pr} that the DeepSORT baseline is largely invariant to this parameter, as it does not report its internal forecasts.
Reporting these estimates, whether directly (corresponding to `DeepSORT+Forecast') or with our approach (corresponding to `Our Pipeline'), highlights the impact of the parameter.
This behaviour results in a precision-recall `curvelet' which shows that by increasing $N_{age}$, we can trade-off the precision and recall for invisible people detection. The difficulty of this task can be highlighted by the trend that increasing $N_{age}$ hardly increases recall beyond a point but instead decreases precision dramatically because of the introduction of many false positive boxes in the scene.
We use the number of frames as a surrogate for uncertainty, as we find that this correlates well with the uncertainty estimated by the Kalman Filter, as shown in Figure 4 in the main paper. 

 We use a hyperparameter $f_{process}$ to scale the process noise covariance (refer Section 3.3 in the main paper). We additionally scale the observation noise covariance by $f_{observation}$ to account for the removal of default scaling by height of \cite{wojke2017simple}. In our algorithm, we use $f_{process}=900$ and $f_{observation}=600$.

\section{MOTA-occluded}
\label{sec:supp_mota}

In the main paper, we report results using the IDF1 tracking metric in addition to the detection F1 metric.
Here, we supplement these results with the MOTA (Multi-Object Tracking Accuracy metric~\cite{bernardin2008evaluating}.
To do this, we follow the strategy in the main paper: We do not penalize tracks that match to visible people, but we reward only tracks that match to occluded people.
For MOTA, we count detections matching to occluded groundtruth as true positives (TP), unmatched detections as false positives (FP), and unmatched groundtruth as false negatives (FN), and only count ID-switches (IDS) for tracks corresponding to occluded groundtruth.
Perhaps surprisingly, we find in Table~\ref{tab:tracking} that the MOTA metric is negative for all ablations.
To better understand this, we note that MOTA is a simple combination of TP, FP, and identity switches (IDS), divided by the total number of groundtruth boxes (GT):
\begin{align*}
\text{MOTA} = 1 - \frac{\sum_t \text{FP}_t + \text{FN}_t + \text{IDS}_t}{\sum_t \text{GT}_t}
\end{align*}
Thus, a method which simply reports no tracks will achieve a MOTA of $0$ (as $\text{FP}=0, \text{FN}=\text{GT}, \text{IDS}=0$), seemingly outperforming all approaches in Table~\ref{tab:tracking}.
This suggests MOTA penalizes methods for even \textit{trying} to detect occluded people.
In general, if a tracker produces more false positives than true positives, MOTA will always be negative!
This indicates that MOTA is not an appropriate metric for challenging tasks, such as detecting occluded people. 
 
\begin{table}[t]
\centering
    \begin{tabular}{lcc}
    \toprule[1.5pt]
    & \multicolumn{2}{c}{MOTA} \\
    \cmidrule(lr){2-3}
    & Occl & All \\
    \midrule[.5pt]
    DeepSORT &
         \textcolor{gray}{-11.9} & 49.4\\
    + Forecast &
         \textcolor{gray}{-85.7} & 42.0\\
    + Egomotion &
         \textcolor{gray}{-72.1} & 44.6\\
    + Freespace &
          \textcolor{gray}{-35.2} & 48.1\\
    + Dep. noise &
           \textcolor{gray}{-31.5} & 48.5\\
    \bottomrule[1.5pt]
    \end{tabular}
    \caption{Analysis of MOTA-occluded for the MOT-17 train ablation experiments. Note that MOTA is not useful for distinguishing trackers for difficult tasks, as it leads to negative values (while an approach which reports no detections would achieve MOTA of 0). %
    \vspace{-4mm}}
    \label{tab:tracking}
\end{table}
\section{Human Vision Experiment}
\label{sec:supp_human_vision}
In the main paper, we briefly described our human vision experiment to understand the challenges in detecting occluded people, and to motivate our evaluation and probabilistic approach.
We provide further details here.
We ask 10 in-house annotators to label fully occluded people in the MOT-17 \cite{milan2016mot16} training set.
To focus annotation effort on occluded people, we sampled track segments (1) containing at least 10 contiguous occluded frames, preceded by (2) 10 frames where the person is visible (and at least one where the person has $>70$\% visibility). Additionally, we avoid annotating small people ($<20$ pixels on either side), and limit the number of total frames in a segment to 50.

Annotators labeled at 10 fps (every 3rd frame in a 30fps video) in a simulated \textit{online} setup. When an annotator is asked to label frame $t$, she has access to past frames (before $t$), but \textit{not} future frames $>t$. Once the annotator submits a label for $t$, she is shown the next frame to label, and is no longer allowed to edit the annotation for frame $t$.

Overall, 10 people labeled a total of 113 tracks, 46 of which were unique. This resulted in a total of 991 annotated boxes. Our key finding was that even for complete occlusions (less than 10\% visibility), annotators still agreed to a fair extent (60\% IoU-agreement), making the problem harder than localizing visible people, but still feasible for humans.
To account for these observations, we evaluate with our invisible-people detection metric at an IoU of 0.5.

\section{PANDA and MOT-20}
\label{sec:supp_panda_mot20}

\newstuff{We first discuss the quality of visibility labels in PANDA followed by the criteria we follow for disabling the depth and freespace reasoning in our method for a subset of videos in PANDA \cite{wang2020panda} and MOT-20 \cite{dendorfer2020mot20}. }

\begin{figure}[h] %
  \centering
    \includegraphics[width=\textwidth]{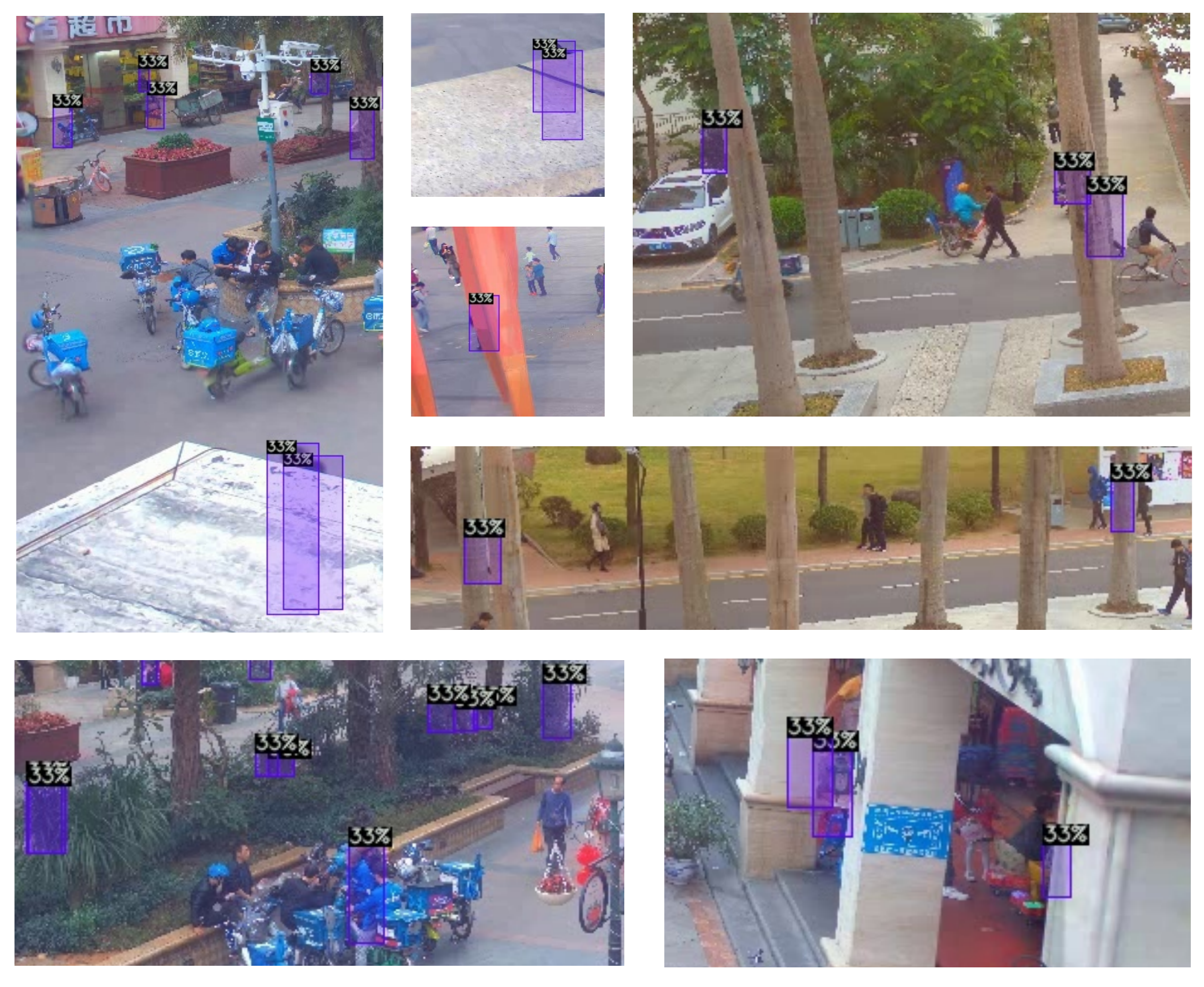}
  \caption{`Heavy occlusion' or 33\% visibility labels in PANDA are closer to the $<10$\% visibility labels in the MOT-17 and MOT-20 datasets. For this reason, we set the visibility threshold in the PANDA dataset to 33\%.
  \vspace{-3mm}
  }
  \label{fig:pandavis}
\end{figure}

\newstuff{PANDA classifies the visibility of people into 4 discrete classes -- `without occlusion', `partial occlusion', `heavy occlusion' and `disappearing'. According to the dataset authors, these correspond to 100\%, 66\%, 33\% and 0\% visibility labels on a continuous 0-100 scale. On qualitative inspection, we find that most 33\% visible people in PANDA are fully-occluded (by our definition of $<$ 10\% visibility). Though the visibility annotation protocol is not detailed in the paper, we hypothesize that this anomaly exists because only those people are marked with 0\% visibility which strictly have 0 visible pixels. Some examples are shown in Figure \ref{fig:pandavis}. Owing to this, we set the threshold of calling a person invisible in the PANDA dataset as 33\% visibility.

Some sequences in PANDA and MOT-20 are top-down view videos where occlusions are unlikely to occur.
In such sequences, we revert to using the standard DeepSORT tracker.
For MOT-20, we disable our method on two sequences captured from a camera mounted at a high height based on visual inspection.
For the PANDA dataset, which specifies the building floor on which the camera is mounted, we use DeepSORT for cameras mounted on or above the 8th floor.
We note that this decision can be easily made in the real world by practitioners based on the height of the camera.
}
\section{Pseudo-code}
\label{sec:supp_pseudocode}

In Algorithm \ref{algo:kf}, we present the pseudocode of our approach for detecting occluded people. Execution starts from the \textit{main()} function.

\begin{algorithm}[h]
\SetAlgoLined
\KwData Detections $\d$ in current frame, $f_i \in \mathcal{F}$, the set of all frames \\
\KwResult{Set of active tracks, $\t = \{t_1, \dots, t_k\}$ s.t. $t_j \in \{\t_{occluded}, \t_{visible}\}$}
\vspace{4mm}
\SetKwProg{Def}{def}{:}{}
\Def{update()}{
    X, Y1, Y2, Z = \textit{match()}\;
    Update the tracks with the KF Update step for all pairs in X\;
    Initialise new tracks for Z\;
    Increase age of all tracks in Y1\;
    Add Y2 to $\t_{occluded}$\;
    }
    \caption{Invisible-people Kalman Tracker}
    \label{algo:kf}
\end{algorithm}
\begin{algorithm}
\SetKwProg{Def}{def}{:}{}
\Def{predict()}{
    Find warp marix $W$ between current and past frame\;
    \For{all active tracks}{
        Warp the mean of current tracker state with the warp matrix\;
        Assume a Constant Velocity Model\;
        If track is occluded, assume no velocity for $a$ and $h$\;
        Else, assume constant velocity for $a$ and $h$\;
        Assume temporal process noise for all state variables (e.g., process noise $f\frac{\epsilon_X}{Z}$ for $x$)\;
        Carry out the KF Predict step to get a new state from the warped state\;
    }
    }
    
\vspace{3mm}
\Def{match()}{
    Compare forecasted depth, $z_f$ with horizon depth, $z_o$\;
    If $z_f < \alpha_{supp}z_o$, keep track in $\t_{visible}$ but don't output\;
    Else, trigger occluded state logic by adding track to $\t_{occluded}$\;
    Bipartite-match detections to active tracks to based on last-known appearance\;
    Match unclaimed visible tracks to unclaimed detections using IoU\;
    Let X be matched tracks and detection\;
    Let Y be unclaimed tracks\;
    Let Z be unclaimed detections\;
    Separate Y into visible (Y1) and occluded (Y2) tracks\;
    \For{all tracks in Y2}{
    If $z_f < \alpha_{delete}z_o$, delete track\;}
    Return X,Y1,Y2,Z\;
     }
\vspace{3mm}
\Def{main()}{
    \For{every incoming frame}{
        predict new states for all tracks using \textit{predict()}\;
        update all tracks with detections from the current frame using \textit{update()}\;
        output all active tracks that are either currently occluded or visible\;
    }
    }

\end{algorithm}

{\small
\bibliographystyle{ieee_fullname}
\bibliography{egbib}
}

\end{document}